\documentclass{article}

\PassOptionsToPackage{numbers, compress}{natbib}

\usepackage{iclr2025_conference,times}

\usepackage[utf8]{inputenc} 
\usepackage[T1]{fontenc}   
\usepackage{hyperref}       
\usepackage{url}            
\usepackage{booktabs}       
\usepackage{amsfonts}       
\usepackage{nicefrac}       
\usepackage{microtype}      
\usepackage{xcolor}        
\usepackage{graphicx}

\title{Decoupling the components of geometric understanding in Vision Language Models}

\author{
    \textbf{Eliza Kosoy$^{1*}$, Annya Dahmani$^{2*}$, Andrew K. Lampinen$^{1}$, Iulia M. Comșa$^{1}$,} \\
    \textbf{Soojin Jeong$^{1}$, Ishita Dasgupta$^{1*}$, Kelsey Allen$^{1}$}\thanks{Equal contribution} \\
    \textsuperscript{1}Google DeepMind, Mountain View, CA, USA \\
    \textsuperscript{2}UC Berkeley, Berkeley, CA, USA \\
    \texttt{adahmani@berkeley.edu}, \texttt {\{elko, lampinen, iuliacomsa}, \\ \texttt {soojinj, idg, krallen\}@google.com}
}

\iclrpreprintcopy

\begin{document}

\maketitle

\begin{abstract}
  Understanding geometry relies heavily on vision. In this work, we evaluate whether state-of-the-art vision language models (VLMs) can understand simple geometric concepts.
  We use a paradigm from cognitive science that isolates visual understanding of simple geometry from the many other capabilities it is often conflated with such as reasoning and world knowledge. We compare model performance with human adults from the USA, as well as with prior research on human adults without formal education from an Amazonian indigenous group. We find that VLMs consistently underperform both groups of human adults, although they succeed with some concepts more than others. We also find that VLM geometric understanding is more brittle than human understanding, and is not robust when tasks require mental rotation. This work highlights interesting differences in the origin of geometric understanding in humans and machines -- e.g. from printed materials used in formal education vs. interactions with the physical world or a combination of the two -- and a small step toward understanding these differences.
\end{abstract}

\section{Introduction}

A substantial amount of recent work has tried to understand how well vision-language models understand visual concepts \citep{patraucean2024perception,mouselinos2024beyond,kamoi2024visonlyqa,ollikka2025comparison}. Here, we focus on geometry. Geometry is complex and relies heavily on visual understanding in humans (though AI approaches often sidestep the visual component, e.g. \citealp{trinh2024solving}). However, the visual-understanding component of geometry task performance is often entangled with other capabilities -- for example, geometric stimuli often occur as a figure in a math textbook, where "understanding" the figure might include the ability to read the labels in the image (OCR), to recognize the mathematical concepts conveyed (world knowledge), and to use this information to then answer some math question (multi-step reasoning). Several geometry or geometry-adjacent benchmarks \citep[e.g.][]{kazemi2023geomverse} conflate these different capabilities. \citet{kamoi2024visonlyqa} attempt to remove reasoning components, but still incorporate some aspects of reading, understanding formal mathematics, etc.

In this work, we are interested in isolating whether VLMs understand basic geometric concepts visually -- separately from knowledge that might arise from a formal education in mathematics. To this end, we build on a long history of work in cognitive science that attempts to assess the origins of geometric understanding \citep{piaget1948child,spelke2011natural,spelke2012core}. Specifically, we adapt stimuli from \citet{dehaene2006core}, who investigated the geometric knowledge of the Munduruku, an indigenous Amazonian group, to understand whether fundamental geometric concepts are present across cultural and educational differences. We also adapt their experimental design where rather than asking participants to verbalize specific geometric concepts, participants are presented with 6 images, 5 of which satisfy some geometric concept (e.g. symmetry, or connectedness) and one that violates this concept. Participants are then asked to select the odd one out.

We also test adult participants from the USA on the same concepts using Prolific \citep{palan2018prolific}, and compare to the Munduruku's performance as reported in \citet{dehaene2006core}. Our overall findings show that even adult humans do not uniformly agree at some of these tasks, and that there is an overall trend that US-adults perform most consistently, followed by the Munduruku, followed by the VLMs. Here, we focus our investigations on a few concepts that showed the maximum deviation or variability in VLM performance. Based on these exploratory analyses, we found that there was still another capability entangled with geometric understanding implicit in the stimuli -- that of mental rotation.

Many experiments in cognitive science show that humans engage in mental rotation to understand visual stimuli \citep[e.g.][]{shepard1971mental}. Several of the geometry stimuli from \citet{dehaene2006core} implicitly assume this capability \ref{fig:rotation}A. This assumption seems valid for the Munduruku -- on a different set of stimuli also from \citet{dehaene2006core}, they find "Performance
with allocentric, egocentric, and rotated maps did
not differ and was always significantly
above chance, suggesting that participants either
extracted the geometrical relationships directly
or performed a mental rotation so as to align the
map with the environment." However, it remains to be seen if this assumption is valid for VLMs; recent results suggest that VLMs' perception is less robust to rotation \citep{ollikka2025comparison}. We therefore generated rotation-controlled versions for 4 of the stimulus sets from the broader geometry set, and conducted experiments on both US adults and VLMs. In these stimulus sets, all 6 images are presented with the same global angle of orientation, therefore eliminating the need for mental rotation. We found that human performance was relatively unaffected (both by the need to realign stimuli with mental rotation, as well as the final rotation angle of stimuli presented to the participants), while VLM performance varied dramatically across rotation angle, and was consistently lowest in the condition that required mental rotation.

\begin{figure}
    \centering
    \includegraphics[width=1.0\linewidth]{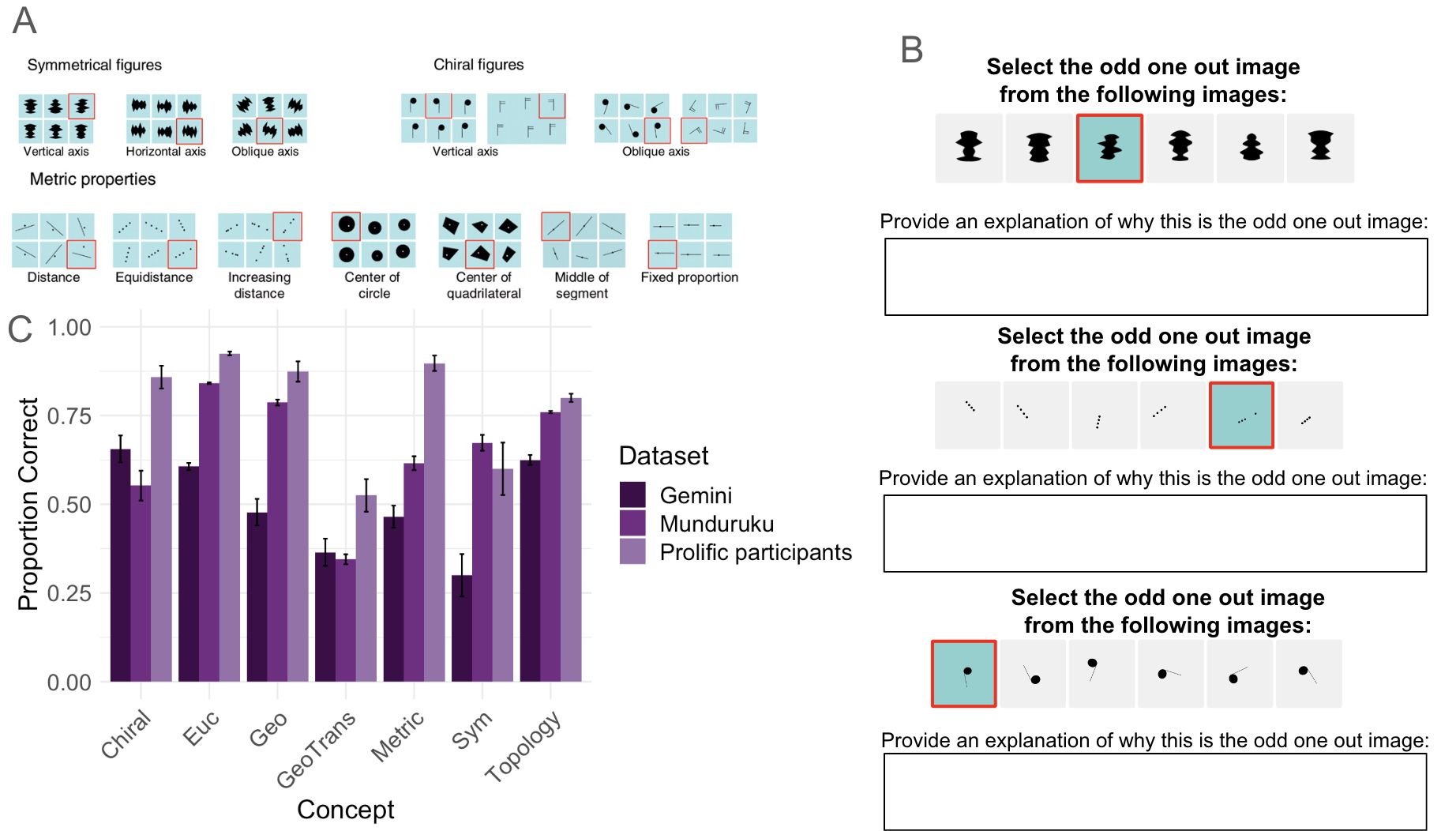}
    \caption{\textbf{Overview.} A. Adapted from \cite{dehaene2006core} to illustrate the geometric concepts we explore in this work. For the full set of stimuli refer to Appendix \ref{fig:all_stim}. B. An example of the experimental protocol. This is what human participants saw, VLMs received a similar prompt. See main text for details. C. Overview of results; see main text for discussion.}
    \label{fig:overview}
\end{figure}

\section{Methods}
\subsection{Adapting stimuli}
The geometric stimuli were copied from \cite{dehaene2006core} and \cite{franco1977hemisphere} and created in the free software program Inkscape by tracing the stimuli from the images in the original papers. The images were all black and white. To investigate the effects of mental rotation, we also generated rotation-controlled versions of the original stimuli. The stimuli represent various forms of geometrical shapes from 6 concepts, these were: points, lines, parallelism, 
figures, congruence, and symmetry. See Figure 3 for actual examples. Within each category there were various numbers of trials. These were originally selected by \cite{dehaene2006core} to test intuitive comprehension of basic geometrical concepts. For each trial, there were six images, one of which violated the core concept. 

\subsection{VLM experiments}
We evaluated the Gemini Pro 1.5 model \citep{team2024gemini} with default parameters. We used the following prompt: \texttt{I will give you 6 images, you have to choose which one is the odd one out. Respond with the index (between 1 and 6) of the odd one out image, followed by the word `EXPLANATION:',followed by an explanation of why this is the odd one out image.}. We ran this 20 times per trial, randomizing the order of the images presented to the model each time. Error bars are calculated over these 20 runs. We used the explanations to debug our process, but did not use them in our final analyses.

\subsection{Human experiments}

We recruited \textit{N} = 15 U.S. adults (mean age = 39.53, $\sigma=9.56$) on Prolific \cite{palan2018prolific} to complete a similar task as the model. Our selection criteria targeted participants who were from the United States, between the ages of 18-99, fluent in English, and had completed at least 50 previous Prolific studies.

Participants were instructed to choose the "odd one out" image, defined to participants as the image different from the others. Mirroring \citet{dehaene2006core}, participants completed two training trials: one based on orientation and the other on color. Feedback was provided for incorrect responses, and participants repeated the trials until correct. Participants then completed the remaining stimuli (same as what the model saw), selecting the "odd one out" image from 43 randomized trials from\citep{dehaene2006core}. Additionally, we included 10 new stimuli with varying degree permutations (0, 45, and 90 degrees) across specific categories (Metric Properties: Equidistance and Middle, Chiral Figures: Vertical Axis 1 and 2) to explore changes in performance (example of Chiral Figures Vertical Axis 1 rotation degrees in Fig \ref{fig:rotation}. In total, participants completed 53 vignettes without feedback and provided explanations for their choices, which are currently under analysis.

\section{Results}

\subsection{Overall geometric understanding}

Note that the odd one out task is under-specified -- there is not a definitively correct answer. In principle, participants could choose any object since they all have unique features. However, in practice, all choices tended to cluster along the intended dimensions, with all conditions yielding performance well above 16.7\% chance, revealing overall sensitivity to these geometric concepts. Among the groups, both Prolific adults and the VLM have access to data that goes into a formal education in basic geometry while the Munduruku does not. As represented in Fig \ref{fig:overview}, we find the following:
\begin{itemize}
    \item Prolific adults do not score perfectly, and there is large variation across categories. 
    \item Munduruku adults score lower than Prolific adults. This possibly suggests that formal education biases us towards categorizing along geometric features, but requires more experimentation (e.g. controlling stimuli differences better) to make strong conclusions.
    \item Gemini scores lowest on most stimuli. This difference could be driven by both the model's inability to recognize the geometric concepts embedded in the image, or by preferring to categorize by other dimensions (in the under-specified odd one out setting).
\end{itemize}

Performance is more similar between the Munduruku and the Prolific adults than the VLM and Prolific adults. A factor that sets the humans apart from the VLM is their exposure to physical or real-world interactive experience. In the next section, we explore one way that this might affect performance on our task.

\subsection{Understanding the effect of mental rotation}

\begin{figure}
    \centering
    \includegraphics[width=.95\linewidth]{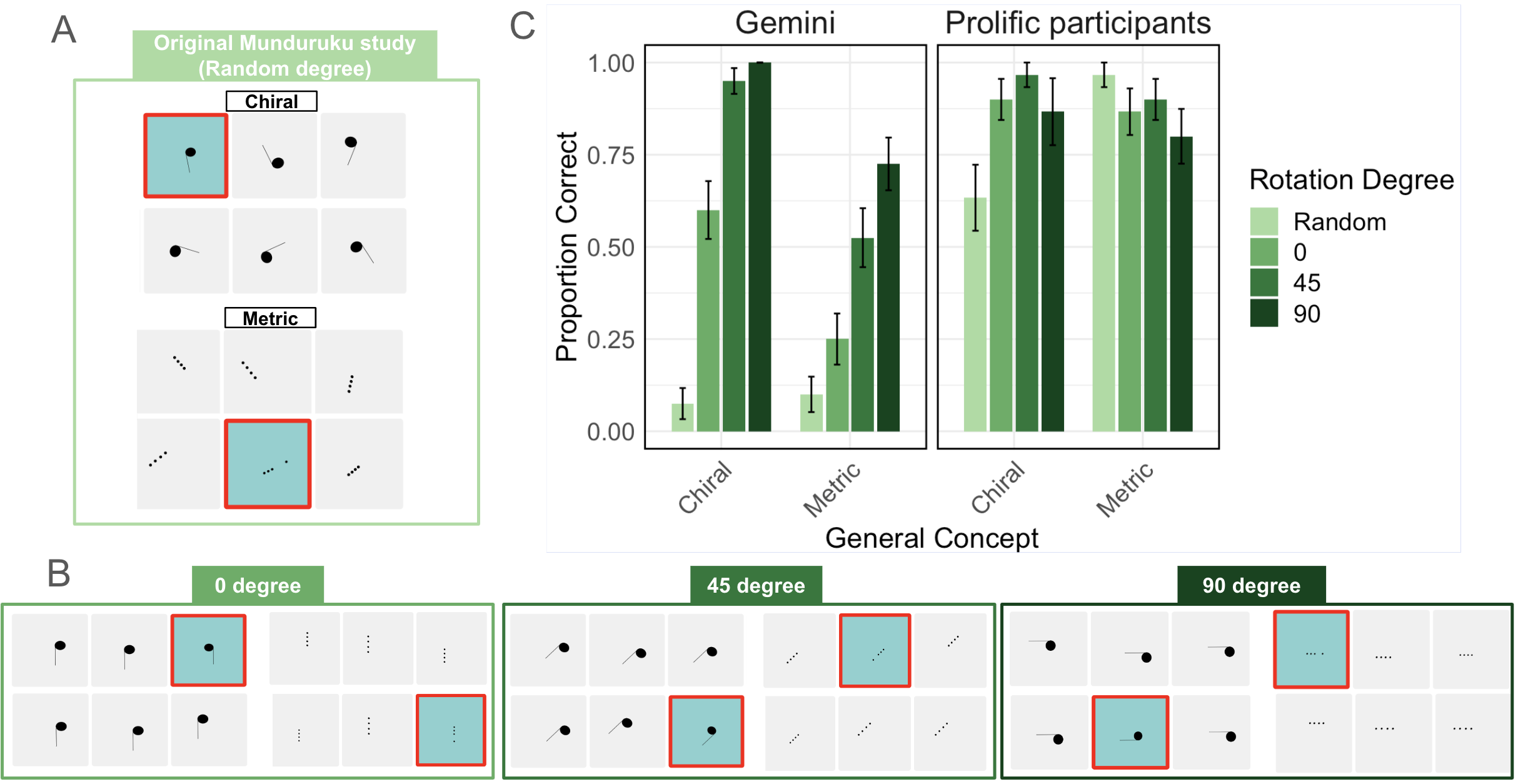}
    \caption{\textbf{Mental rotation.} A. Stimuli inspired by the original Munduruku study aimed to measure sensitivity to chirality and metric equidistance, that implicitly assumes mental rotation. B. Rotation controlled version of the chirality and metric. C. Results on humans and Gemini on rotation controlled stimuli. The Munduruku were only evaluated for "random", and achieve approximately 60$\%$ in both the chiral and metric categories.}
    \label{fig:rotation}
\end{figure}

 In the stimulus sets from \citep{dehaene2006core}, several assume capacity for mental rotation (e.g. Fig \ref{fig:rotation}A). This capability is implicit in many physical real-world interactions, but not necessarily in formal education materials about geometry, which tend to be presented in canonical orientations. We create versions of our stimuli that control for the need to use mental rotation (Fig \ref{fig:rotation}B), and find that while humans are not significantly affected, Gemini shows substantial variation in behavior across different rotation angles. The model performs nearly as well as humans in the 90 degree condition, but falls off when the stimuli are oriented away from this angle, and shows exceptionally poor performance on the cases where examples appear at random rotation angles (Fig \ref{fig:rotation}C). Thus, the model is not as robust as humans to comparisons at novel angles or to tasks that implicitly require rotating stimuli by different angles to compare them.

\section{Discussion}
This work offers a first step towards exploring the intricacies of geometric understanding in artificial models. Our results highlight the importance of separating capabilities to be tested, and not assuming related capabilities (such as mental rotation) simply because we can assume them in humans. Our work also raises questions about the different origins of geometric understanding in humans and machines, and the role of formal education. VLMs learn about geometry from the millions of geometric figures in math textbooks online, while the Munduruku have not seen printed material but have acquired those same concepts by exposure and interaction with the physical world. The stark gap in VLM and human capability to perform mental rotation is a symptom of this difference. An interesting direction of future work is to understand more qualitatively the behavior of US-adults with formal education who have access to both printed material as well as the physical world -- and how their geometry understanding combines these sources of learning. Future work can also look at different VLMs, to evaluate differences in their behaviors based on possible differences in their training protocols.

\subsection*{Acknowledgements}

We thank Maria Eckstein and Mike Mozer for helpful comments and suggestions.

\bibliography{iclr2025_conference}
\bibliographystyle{iclr2025_conference}

\clearpage
\appendix

\section{Appendix / supplemental material}

\begin{figure}[ht!]
    \centering
    \includegraphics[width=1\linewidth]{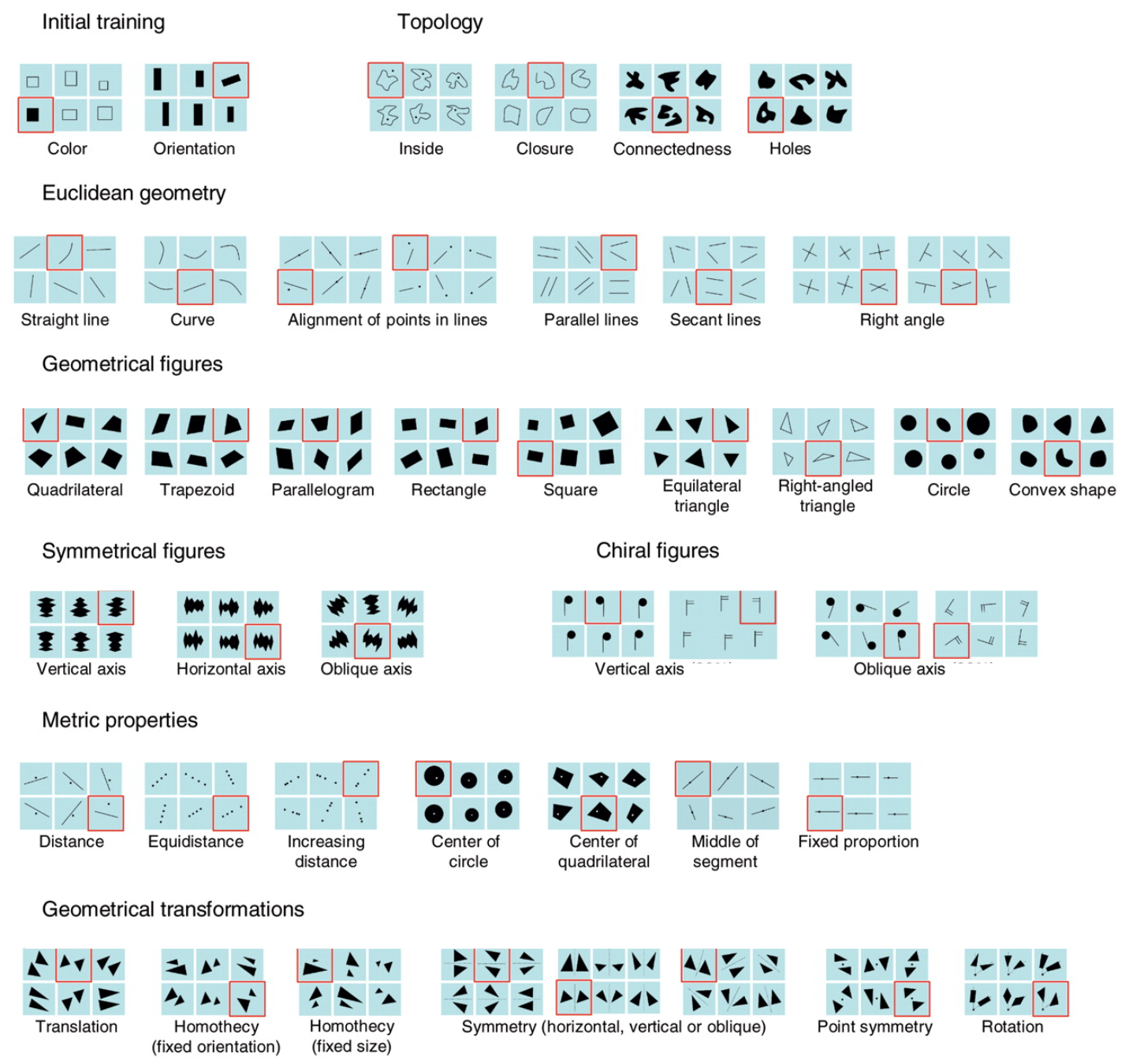}
    \caption{The full set of stimuli adapted from \cite{dehaene2006core} of the geometric concepts we explore in this work.}
    \label{fig:all_stim}
\end{figure}

\begin{figure}[ht!]
    \centering
    \includegraphics[width=\linewidth]{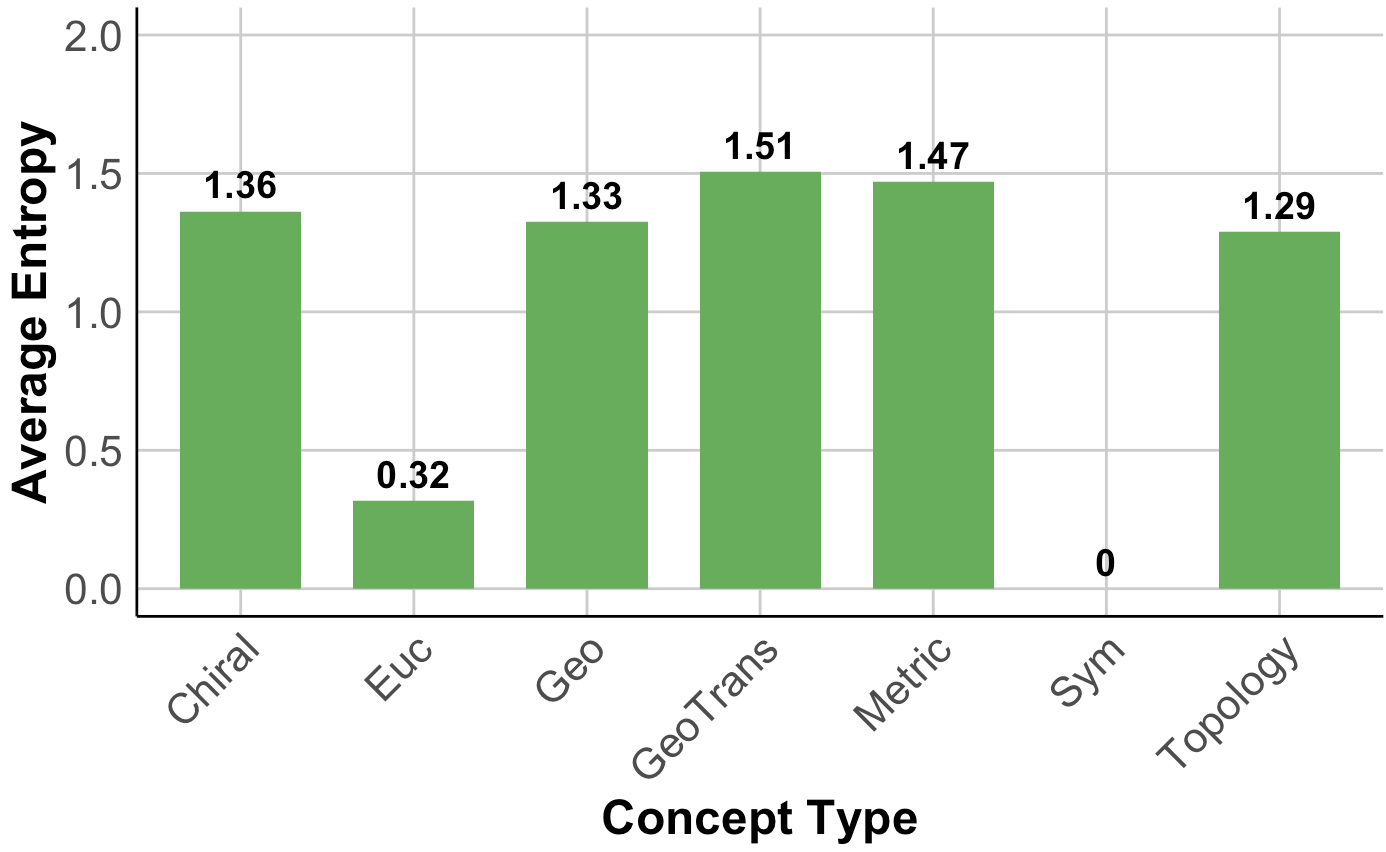}
    \caption{\textbf{Entropy of Gemini choices across categories.}}
    \label{fig:entropy}
    
\end{figure}

\begin{figure}[ht!]
    \centering
    \includegraphics[width=1.0\linewidth]{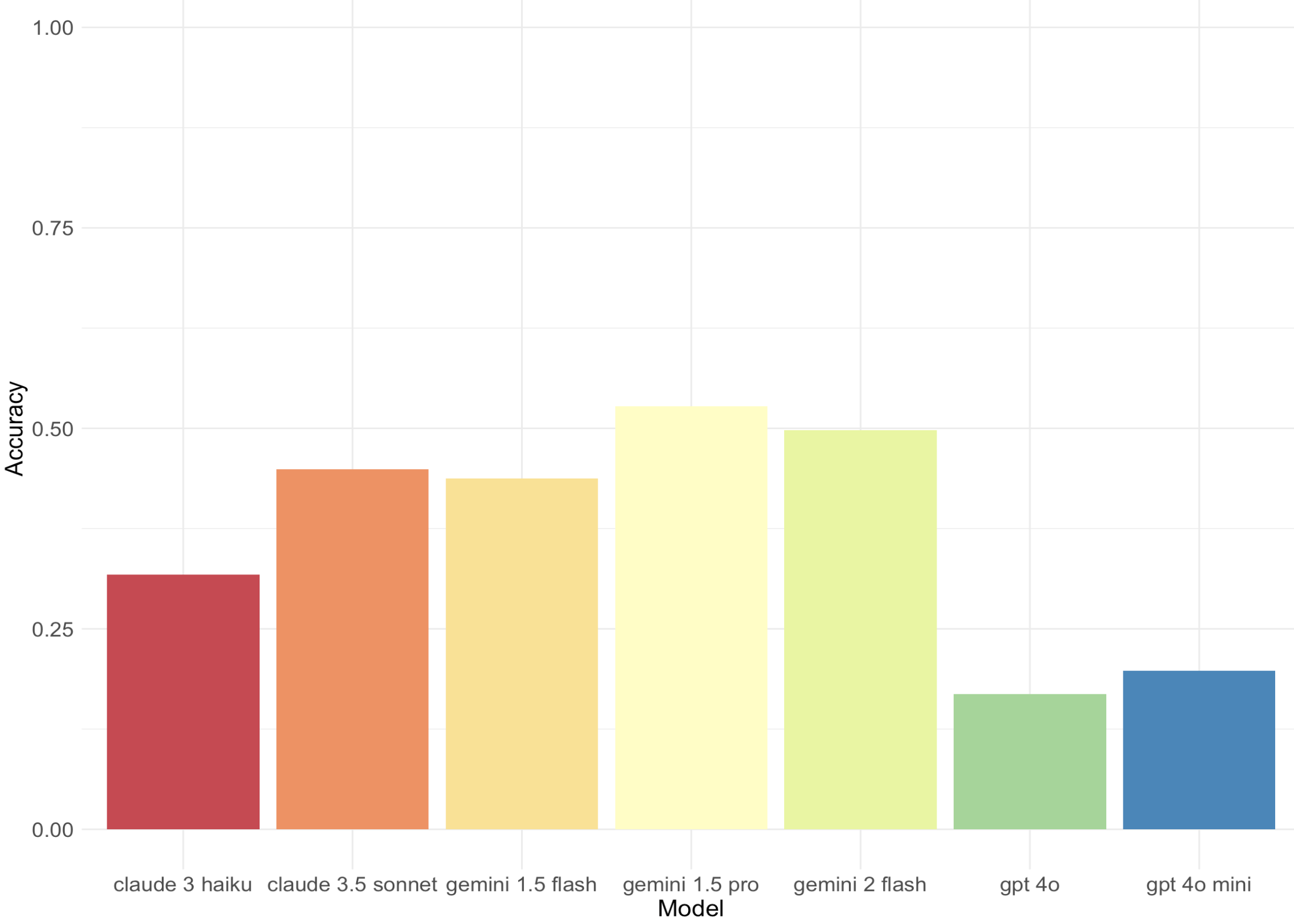}
    \caption{\textbf{Accuracy across 7 vision-language models  in geometric concept task.}}
    \label{fig:sall_model_accuracy}
\end{figure}

\begin{figure}[ht!]
    \centering
    \includegraphics[width=1.2\linewidth]{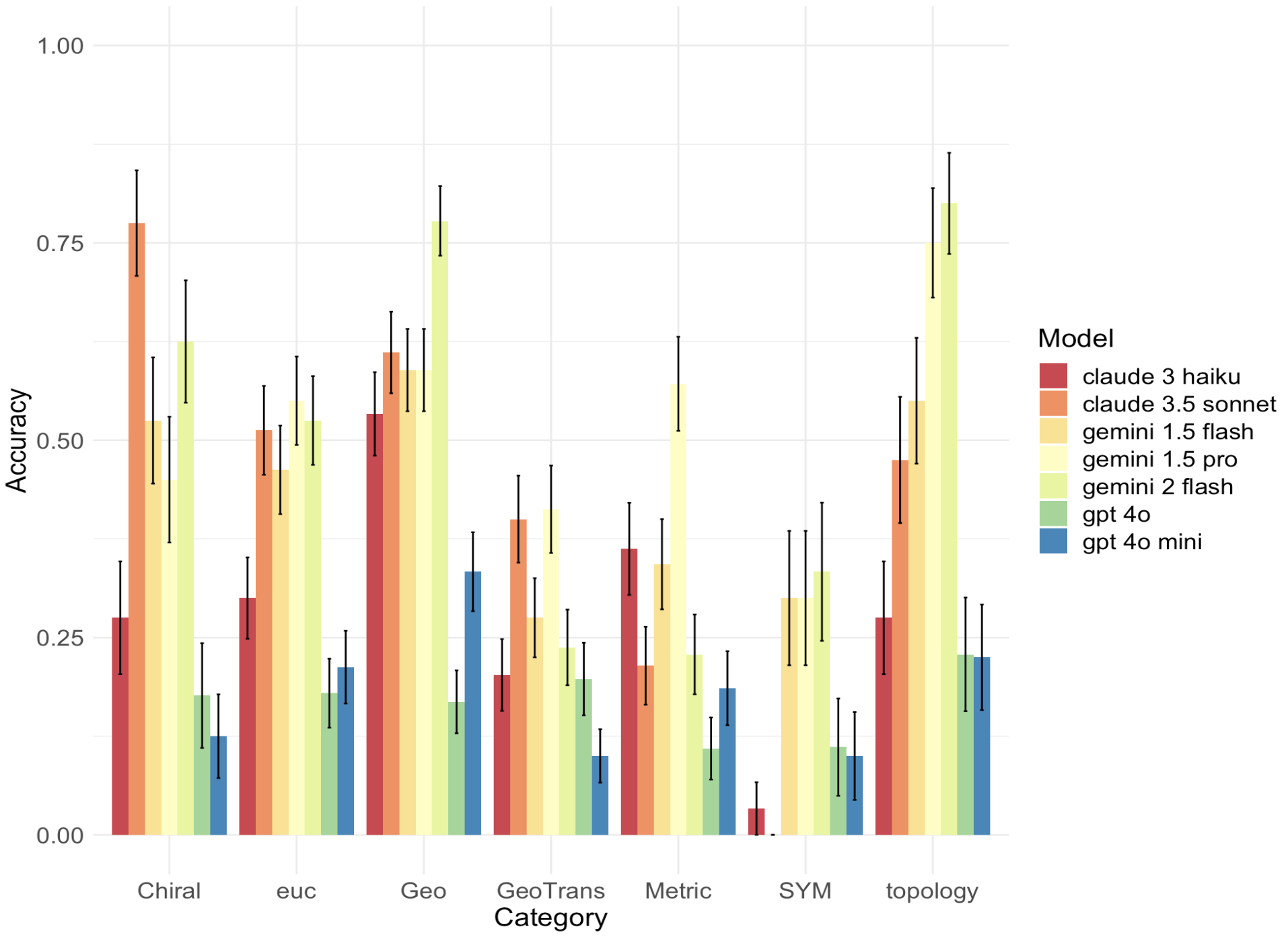}
    \caption{\textbf{Accuracy of all models  in geometric concept task by category.}}
    \label{fig:all_models_category}
    
\end{figure}


\end{document}